%% file: root.tex
\documentclass[letterpaper, 10 pt, conference]{ieeeconf}
\usepackage{bm}
\usepackage{color,colortbl}
\usepackage{xcolor}
\usepackage{nicefrac}
\usepackage{caption}
\usepackage{csquotes}  
\usepackage{tabularx}
\usepackage{subcaption}

\usepackage{times}
\usepackage{epsfig}
\usepackage{graphicx}
\usepackage{amsmath}
\usepackage{amssymb}
\usepackage{gensymb}
\usepackage{float}
\usepackage{mathtools}
\usepackage{dblfloatfix}

\usepackage{multirow}
\usepackage{array}
\usepackage{booktabs}
\usepackage[export]{adjustbox}
\usepackage{xfrac}
\usepackage{stmaryrd}

\usepackage{comment}
\usepackage{multirow}
\usepackage{array}
\usepackage{booktabs}
\usepackage{xspace}
\usepackage{algorithm}
\usepackage{algpseudocode}

\makeatletter
\let\NAT@parse\undefined
\makeatother

\usepackage[pagebackref=true,breaklinks=true,letterpaper=true,colorlinks,bookmarks=false]{hyperref}

\definecolor{darkred}{rgb}{0.55, 0.0, 0.0}
\definecolor{darkblue}{rgb}{0.0, 0.0, 0.55}

\makeatletter
\DeclareRobustCommand\onedot{\futurelet\@let@token\@onedot}
\def\@onedot{\ifx\@let@token.\else.\null\fi\xspace}

\def\eg{\emph{e.g}\onedot} 
\def\ie{\emph{i.e}\onedot}

\makeatother

\begin{document}
%
\title{Learning from Spatio-temporal Correlation for Semi-Supervised LiDAR Semantic Segmentation}
%
%
%
\author{Seungho Lee\\
Yonsei University\\
{\tt\small seungholee@yonsei.ac.kr}
\and
Hwijeong Lee \\
KAIST \\
{\tt\small hjlee0612@kaist.ac.kr}
\and 
Hyunjung Shim \\
KAIST\\
{\tt\small kateshim@kaist.ac.kr}}
\markboth{}
{} 

%



\maketitle

\begin{abstract}
\input{sections/0_abstract}
\end{abstract}

\input{sections/1_intro}

\input{sections/2_related}

\input{sections/3_method}

\input{sections/4_experiments}

\input{sections/5_conclusion}


\noindent \textbf{Acknowledgments.} 
This research was supported by the Basic Science Research Program through the National Research Foundation of Korea (NRF) funded by the MSIP (NRF-2022R1A2C3011154, RS-2023-00219019, RS-2023-00240135) and MOE (NRF-2022R1A6A3A13073319), the IITP grant funded by the Korea government(MSIT) (RS-2019-II190075, Artificial Intelligence Graduate School Program(KAIST)) and KEIT grant funded by the Korea government(MOTIE) (No. 2022-0-00680, 2022-0-01045, 2021-0-02068, Artificial Intelligence Innovation Hub) and Samsung Electronics Co., Ltd (IO230508-06190-01).






%



\bibliographystyle{IEEEtran}
\bibliography{references}


%








\end{document}

%% file: sections/0_abstract.tex
We address the challenges of the semi-supervised LiDAR segmentation (SSLS) problem, particularly in low-budget scenarios. The two main issues in low-budget SSLS are the poor-quality pseudo-labels for unlabeled data, and the performance drops due to the significant imbalance between ground-truth and pseudo-labels. This imbalance leads to a vicious training cycle. To overcome these challenges, we leverage the spatio-temporal prior by recognizing the substantial overlap between temporally adjacent LiDAR scans. We propose a proximity-based label estimation, which generates highly accurate pseudo-labels for unlabeled data by utilizing semantic consistency with adjacent labeled data. Additionally, we enhance this method by progressively expanding the pseudo-labels from the nearest unlabeled scans, which helps significantly reduce errors linked to dynamic classes. Additionally, we employ a dual-branch structure to mitigate performance degradation caused by data imbalance. Experimental results demonstrate remarkable performance in low-budget settings (i.e., $\leq$ 5\%) and meaningful improvements in normal budget settings (i.e., 5 -- 50\%). Finally, our method has achieved new state-of-the-art results on SemanticKITTI and nuScenes in semi-supervised LiDAR segmentation. With only 5\% labeled data, it offers competitive results against fully-supervised counterparts. Moreover, it surpasses the performance of the previous state-of-the-art at 100\% labeled data (75.2\%) using only 20\% of labeled data (76.0\%) on nuScenes. The code is available on \href{https://github.com/halbielee/PLE}{https://github.com/halbielee/PLE}.

%% file: sections/1_intro.tex
\section{Introduction}
\label{sec:intro}

LiDAR segmentation is crucial for autonomous driving, offering robust 3D environment perception and entity identification under various lighting and weather conditions~\cite{roriz2021automotive,li2020lidar}. Current methods employ powerful deep neural networks, requiring extensive labeled data for training~\cite{hu2022sqn,unal2022scribble}. However, annotating LiDAR point clouds is costly and complex, posing a significant challenge.
In contrast, unlabeled LiDAR point clouds are abundantly available at no additional cost. Therefore, we focus on semi-supervised LiDAR segmentation (SSLS), leveraging large unlabeled and small labeled LiDAR point clouds.

\begin{figure}[t]
\begin{center}
\includegraphics[width=\linewidth]{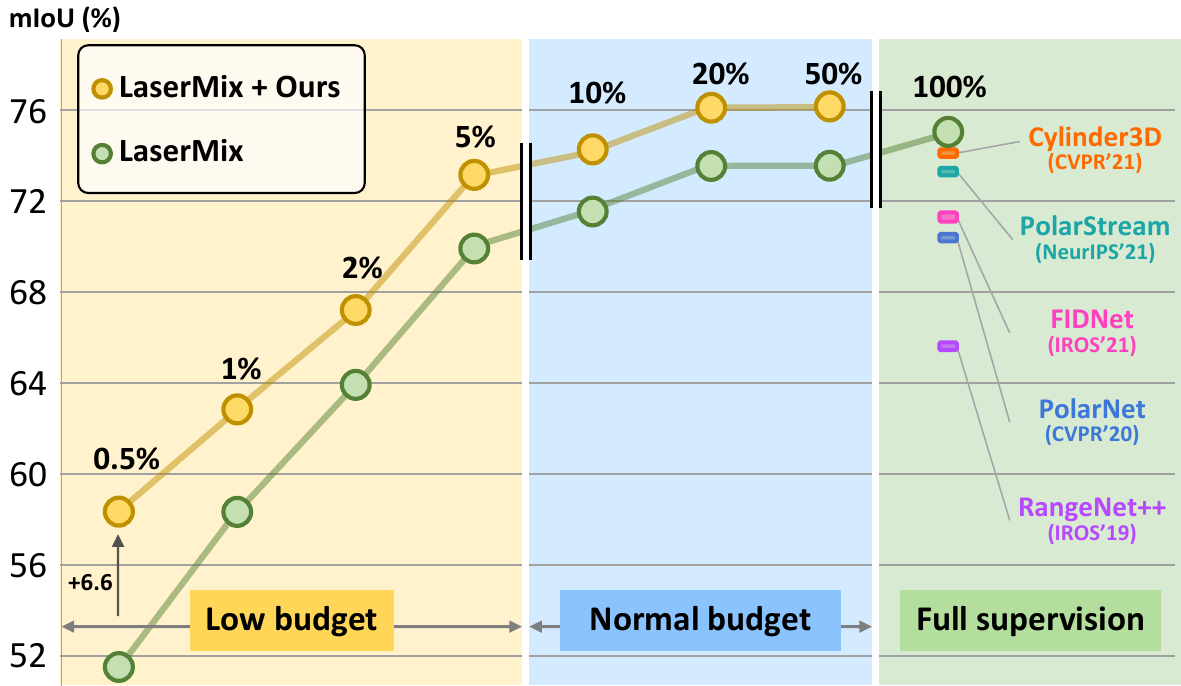}
\end{center}
\caption{\textbf{Segmentation performances (mIoU) across various labeled ratios.} We outperform the state-of-the-art method, LaserMix~\cite{kong2023lasermix}, across all labeled ratios in nuScenes~\cite{caesar2020nuscenes}, with particularly large margins in low-budget settings. Notably, our method with 5$\times$ fewer labels already achieves the performance of full supervision.}
\label{fig:teaser}
\vspace{-1em}
\end{figure}

While semi-supervised semantic segmentation has been extensively studied in 2D images~\cite{wang2022semi,zou2020pseudoseg,luo2020semi,lee2023saliency}, it has been less explored in LiDAR data. Recently, LaserMix~\cite{kong2023lasermix} introduced a new data augmentation method for SSLS, leveraging the spatial prior in LiDAR. By blending laser beams from different LiDAR scans within the same spatial region, LaserMix generates new training samples, boosting performance across various labeled data ratios (\eg, 1\%, 10\%, 20\%, 50\%) and establishing a new state-of-the-art (SoTA) in SSLS.

Despite the impressive achievements, existing studies suffer from a substantial performance drop in low-budget settings (\eg, $\leq$ 5\%). Fig.~\ref{fig:teaser} shows more than 18\%p drop between 0.5\% and 5\%. Reducing labeled data without compromising performance has a significant practical impact, considering the complexity of labeling LiDAR data compared to 2D images.
Therefore, our work focuses on enhancing the practicality of SSLS in low-budget settings.

We identify two main problems of SSLS in low-budget scenarios. (\textcolor{darkblue}{P1}) The pseudo-labels for unlabeled data suffer from poor quality. (\textcolor{darkred}{P2}) A significant imbalance between the amount of labeled and unlabeled data (\eg, 1\% vs 99\%) causes extra performance degradation. To address these problems, (\textcolor{darkblue}{A1}) we propose proximity-based label estimation (PLE) that utilizes the unique spatio-temporal characteristics inherent to LiDAR data to generate high-quality pseudo-labels. Unlike 2D images, LiDAR scans possess temporal continuity, sharing rich semantic information across closely-timed scans~\cite{duong2017reducing}. By aligning unlabeled and labeled scans into the same coordinate, we can accurately assign pseudo-labels to unlabeled LiDAR points using the nearest labeled scan. This strategy can efficiently capture dynamic objects (e.g., vehicles) by progressively extending labels from closer to more distant unlabeled data. Overall, our PLE effectively addresses the challenge of labeling in LiDAR datasets by exploiting their inherent spatio-temporal characteristics.

Additionally, (\textcolor{darkred}{A2}) we employ a dual-branch architecture~\cite{luo2020semi,xie2019intriguing,xie2020adversarial,bermudez2019domain} to mitigate performance degradation caused by data imbalance. Unlike previous methods~\cite{kong2023lasermix}, which rely on a single branch model and struggle with the extreme imbalance between labeled and unlabeled data, our approach offers a robust solution. Specifically, we recognize the popular baseline for SSL, MeanTeacher~\cite{tarvainen2017mean}, falls into a vicious cycle; a poor teacher network produces poor pseudo-labels, which in turn hampers the network training. To break this cycle, we revise the network's final layer into two parallel branches: a labeled branch and an unlabeled branch, each responsible for their respective data. This separation protects the labeled branch from noisy pseudo-labels, leading to stable performance improvement.


In our study, extensive experiments on SSLS benchmarks achievied new SoTA performances with significant improvements on both SemanticKITTI (up to 5.6\%p) and nuScenes (up to 6.6\%p) datasets, particularly in low-budget scenarios. We achieve competitive performance to fully supervised counterparts (100\% labeled) by only utilizing 5\% of full labels and the previous SoTA LaserMix's 100\% labeled data performance using only 20\% of labeled data. Moreover, our experiments demonstrate the effectiveness of PLE for generating accurate pseudo-labels and a dual-branch architecture for improving the training process of the teacher network. Our method proves to be robust and scalable, demonstrating compatibility and substantial performance boosts across different LiDAR representations and backbone networks, underlining the adaptability and potential of our proposed solutions in a variety of settings.

Our key contributions are summarized as follows: (1) We propose a novel approach for generating highly accurate pseudo-labels by utilizing a spatio-temporal prior unique to LiDAR data. (2) We address the performance degradation caused by the data imbalance and break the vicious training cycle in low-budget SSLS with a dual-branch structure. (3) Through extensive experiments, our approach achieves new state-of-the-art results on SemanticKITTI and nuScenes datasets, excelling in low-budget settings, highlighting the effectiveness and practical applicability of our methods.

%% file: sections/2_related.tex
\section{Relatd Work}
\label{sec:formatting}

\subsection{LiDAR Segmentation}

Several studies have explored various representations in LiDAR segmentation. Projection-based methods~\cite{kong2023lasermix, cortinhal2020salsanext, milioto2019rangenet++, xu2020squeezesegv3, zhang2020polarnet, ando2023rangevit, kong2023rethinking} convert LiDAR point clouds into 2D images for processing with 2D convolution. Voxel-based methods~\cite{tang2020searching,zhou2020cylinder3d,thomas2019kpconv} rasterize the LiDAR point clouds into voxels and apply 3D convolution. Multi-view methods~\cite{liong2020amvnet,xu2021rpvnet,zhao2023lif} integrate data from several perspectives. Despite their promising results, these fully-supervised models heavily rely on large-scale, high-quality datasets. Since their labeling cost is prohibitively expensive, recent studies have focused on developing weakly supervised~\cite{unal2022scribble} or semi-supervised methods~\cite{kong2023lasermix,li2023less,ke2020guided} to significantly reduce the annotation cost. In this work, we primarily focus on a semi-supervised learning approach that improves segmentation performance by effectively utilizing readily available unlabeled data.

\subsection{Semi-supervised Segmentation}

Semi-supervised learning has been extensively explored in 2D image segmentation. 
One prevalent strategy is the consistency regularization approach, enforcing consistency between predictions for perturbed data~\cite{zou2020pseudoseg, french2019semi} or networks~\cite{ke2020guided, chen2021semi}. Another approach~\cite{souly2017semi, hung2018adversarial} is to leverage the generative adversarial network (GAN)~\cite{goodfellow2020generative} to provide auxiliary signals for unlabeled data. 
In contrast to 2D segmentation, there is comparatively less research in 3D. Notably, the studies on outdoor scenes have only recently gained attention. GPC~\cite{jiang2021guided} proposes a contrastive learning framework for point clouds but does not differentiate indoor and outdoor scenes. LaserMix~\cite{kong2023lasermix} exploits the spatial prior in outdoor LiDAR scenes and utilizes a mixture of labeled and unlabeled scans along the inclination axis. Recently, LiM3d~\cite{li2023less} introduces the frame downsampling strategy, which determines the most informative keyframes for choosing labeled data. Both LaserMix and LiM3d adopt a pseudo-labeling strategy~\cite{zou2020pseudoseg}, particularly MeanTeacher~\cite{tarvainen2017mean}, repeatedly generating pseudo-labels for unlabeled data. In contrast, we leverage the distinct spatio-temporal prior inherent to LiDAR point clouds and generate highly accurate pseudo-labels for unlabeled data.

\subsection{Temporal information in LiDAR perception}
Temporal information of LiDAR data is commonly used in various fields of LiDAR perception, such as motion prediction~\cite{wu2020motionnet,ye2021tpcn}, LiDAR detection~\cite{huang2020lstm, mccrae20203d, yuan2021temporal, kumar2020real}, sensor calibration~\cite{park2020spatiotemporal}, and scene flow estimation~\cite{liu2019flownet3d,puy2020flot,rishav2020deeplidarflow}. Recently, temporal information has been utilized in self-supervised learning~\cite{nunes2023temporal,wu2023spatiotemporal} and cross-modal 3D understanding~\cite{chen2023clip2scene}. In semantic segmentation, 4D semantic segmentation is actively explored, where multiple scans are taken as input to distinguish between moving and static states of objects~\cite{sun2022efficient}. These studies extract 4D spatio-temporal features by stacking frames~\cite{aygun20214d,kreuzberg20224d}, using k-NN based approaches~\cite{choy20194d,shi2020spsequencenet}, memory-based technique~\cite{schutt2022abstract} or voxel-adjacent neighborbood~\cite{chen2023svqnet}. Recently, LiM3d utilizes temporal information to extract the most informative LiDAR scans. In contrast, our approach focuses on generating accurate pseudo-labels by applyting the spatio-temporal prior.

\begin{figure*}[t!]
\centering
\vspace{1.5mm}
\includegraphics[width=.95\textwidth]{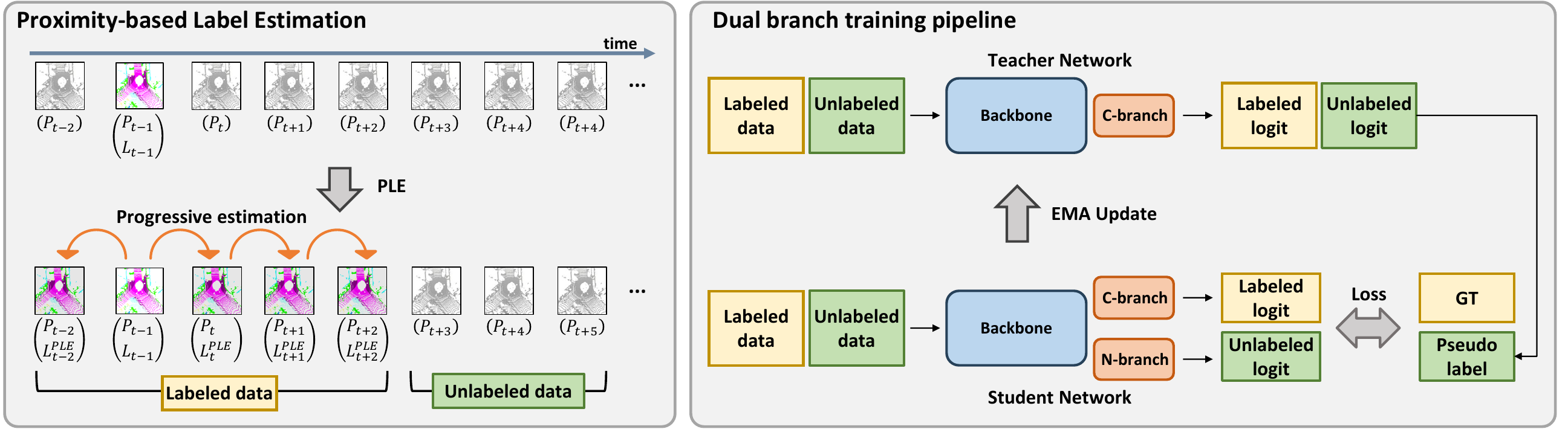}
\caption{\textbf{Overall framework.} 
We generate pseudo-labels for unlabeled scans by leveraging spatio-temporal prior in LiDAR. Unlabeled scans with PLE labels are treated as a labeled set during training. We adopt the MeanTeacher model, where the teacher network generates pseudo-labels for remaining unlabeled data. To mitigate the performance degradation caused by the noisy pseudo-labels, we introduce a dual-branch structure where the labeled and unlabeled data are processed separately.}
 \vspace{-1.5em}
\label{fig:framework}
\end{figure*}

%% file: sections/3_method.tex
\section{Method}

\subsection{SSLS Baseline}\label{section3.1}
Following the previous work~\cite{kong2023lasermix}, we use the MeanTeacher~\cite{tarvainen2017mean}, which involves teacher and student networks. The teacher network is updated from the student network through Exponential Moving Average (EMA) and generates pseudo-labels for unlabeled data. The student network trains on labeled data with ground-truth and unlabeled data with pseudo-labels. We apply cross-entropy and Lov\'asz-softmax losses. We also use the MeanTeacher loss~\cite{tarvainen2017mean} for comparing the two networks' predictions.

\subsection{Proximity-based Label Estimation}\label{section3.2}
Popular benchmark datasets used in 2D segmentation, such as PASCAL VOC~\cite{everingham2015pascal}, COCO~\cite{lin2014microsoft}, and ADE-20k~\cite{zhou2017scene}, do not exhibit spatial or temporal correlations between samples. In contrast, outdoor LiDAR segmentation datasets, such as SemanticKITTI~\cite{behley2019semantickitti} and nuScenes~\cite{caesar2020nuscenes}, present unique characteristics with highly correlated samples in terms of spatial position and sampling time. This is due to the sequential frame-based recording of point clouds as a LiDAR-equipped vehicle moves. Additionally, the high scanning frequency of LiDAR sensors results in substantial overlap between successive scans~\cite{duong2017reducing}. For example, a vehicle moving at 100 kilometers per hour with a LiDAR sensor that has a $\pm 70$ meter range would see an overlap of about 80\% in the 140-meter range across successive frames, significantly due to the rapid data acquisition rate of these sensors.



By leveraging the inherent spatio-temporal prior in LiDAR data, we propose a Proximity-based Label Estimation (PLE) to produce accurate pseudo-labels for unlabeled scans. The key idea is to reference nearby labeled scans based on the spatio-temporal consistency. 
The process of pseudo-label generation consists of three steps. 
(1) \textit{Coordinate transformation}: We transform the coordinate of the referenced labeled scan into the coordinate system of the target unlabeled scan. We utilize pose information, readily available through GPS/IMU~\cite{nunes2023temporal} or ICP~\cite{vizzo2023ral}. 
(2) \textit{Proximate point identification}: We identify the proximate labeled points coinciding with (or closest to) the target unlabeled points in the coordinate system. While the coordinate transformation ensures that the two scans share the same coordinate system, points from the two scans do not exhibit an exact spatial match (\ie, coinciding at the same position). Therefore, we identify the closest labeled counterparts to the unlabeled points. To reduce the searching cost, we build a KD tree~\cite{bentley1975multidimensional} for all transformed labeled points to efficiently compute the distances between target unlabeled points. 
(3) \textit{Label assignment}: We transfer the label of the selected labeled point to the unlabeled point as a pseudo-label. We refer to this pseudo-label as a PLE label to distinguish it from those generated by the teacher network. 

The original PLE is effective in handling static scenes since only global transformation is considered in coordinate transformation. Therefore, it has limitations in allocating labels for dynamic objects, especially when the time interval between ground truth and unlabeled data is sufficiently far. To address this performance limit of dynamic objects, we devise \textit{a progressive version of PLE}. In the original PLE, we only reference ground-truth of the labeled data, regardless of the temporal distance between scans. The progressive PLE addresses this by sequentially generating labels, starting from unlabeled scans closest to labeled ones and progressively moving to more distant scans, as illustrated in Fig.~\ref{fig:framework}. This incremental approach significantly reduces the effective time interval between the labeled and unlabeled data, thereby significantly improving accuracies for dynamic objects.

Despite the simplicity, our PLE method efficiently assigns accurate pseudo-labels to unlabeled data. Since these pseudo-labels from PLE are highly accurate compared to those generated by the teacher network during the training process (Fig.~\ref{fig:ablation}), it significantly improves the training of semi-supervised models. Also, PLE can seamlessly integrate into existing frameworks as it generates pseudo-labels independently of the training process.

\input{tables/benchmark}

\subsection{Dual-branch for low-budget scenario}\label{section3.3}

In the MeanTeacher training process, the student network learns from both ground-truth of labeled data and the pseudo-labels generated by the teacher network for unlabeled data. However, in low-budget scenarios where a small amount of labeled data is available, there exists a significant imbalance between the amount of labeled and unlabeled data. Consequently, the training of student network is heavily influenced by inaccurate pseudo-labels, leading to degraded performance. This, in turn, results in the generation of inaccurate pseudo-labels, establishing a vicious cycle between the generation of pseudo-labels and the training of the network.

To address this issue, we employ a popular dual-branch architecture~\cite{luo2020semi,xie2019intriguing,xie2020adversarial,bermudez2019domain,ren2015faster}, which is widely utilized to capture rich context from various sources, including different object sizes~\cite{chen2017deeplab}, adversarial examples~\cite{xie2019intriguing,xie2020adversarial}, different domains~\cite{bermudez2019domain} and different tasks~\cite{ren2015faster}. Based on the observation that the latter layer of the network is susceptible to noisy labels~\cite{bai2021understanding,kang2019decoupling,baek2022learning}, as shown in Fig.~\ref{fig:framework}, we employ a shared backbone followed by two branches with identical structures to segregate the noisy pseud-labels. The first branch, called the clean branch (C-branch), learns from accurate labels. The second branch, called the noisy branch (N-branch), learns from noisy pseudo-labels. Since the dual-branch effectively separates the training of clean labels (\ie, ground truth labels and PLE labels) from noisy labels (\ie, pseudo-labels from the teacher network), it prevents the quality degradation of pseudo-labels from the teacher. As a result, the vicious cycle between the training of the network training and noisy pseudo-labels is resolved. This approach is particularly effective in low-budget scenarios. Moreover, it makes the network robust to the confidence threshold that determines the quality of pseudo-labels.

\subsection{Overall Pipeline}\label{section3.4}

Our pipeline consists of PLE and the dual-branch structure. Prior to training, we generate PLE labels for some of the unlabeled data. Specifically, we generate pseudo-labels only for frames within a 1-second timeframe of a labeled sample. This is because we have observed that the precision of PLE labels decreases as the temporal gap from the labeled sample increases, as shown in Fig.~\ref{fig:ablation}~(b).

As detailed in Sec.~\ref{section3.1} and \ref{section3.3}, our approach involves two networks: the student network and the teacher network. During the student network's training, the C-branch is trained using the labeled data and some of the unlabeled data with PLE labels, while the N-branch focuses on the rest of the unlabeled data without PLE labels. The teacher network is updated through EMA and generates the pseudo-labels using its C-branch, with the N-branch remaining unused. 


%% file: tables/benchmark.tex
\begin{table*}[t]
\small

\vspace{1.5mm}
\setlength{\tabcolsep}{4pt}
\centering
\begin{tabular}{>{\raggedright\arraybackslash}c|
p{18.5pt}<{\centering}p{18.5pt}<{\centering}
p{18.5pt}<{\centering}p{18.5pt}<{\centering}
p{18.5pt}<{\centering}p{18.5pt}<{\centering}
p{18.5pt}<{\centering}p{18.5pt}<{\centering}|
p{19pt}<{\centering}p{19pt}<{\centering}
p{19pt}<{\centering}p{19pt}<{\centering}
p{19pt}<{\centering}p{19pt}<{\centering}
p{18.5pt}<{\centering}p{18.5pt}<{\centering}}
\toprule
\multirow{2}{*}{Method} & \multicolumn{8}{c|}{SemanticKITTI~\cite{behley2019semantickitti}} & \multicolumn{8}{c}{nuScenes~\cite{caesar2020nuscenes}} \\
& 0.5\% & 1\% & 2\% & 5\% & 10\% & 20\% & 50\% & 100\% & 0.5\% & 1\% & 2\% & 5\% & 10\% & 20\% & 50\% & 100\% \\ \midrule

\textit{Sup.-only}$^{\dagger}$~\cite{zhou2020cylinder3d}
& 40.0 & 47.1 & 50.3 & 55.1 & 57.6 & 57.6 & 58.5 & 58.5 
& 45.6 & 52.0 & 59.2 & 65.3 & 68.4 & 72.0 & 74.0 & 75.2 \\ \midrule
\footnotesize{MeanTeacher$^{\dagger}$~\cite{tarvainen2017mean}}
& 42.2 & 51.0 & 54.6 & 55.6 & 59.1 & 60.1 & 60.4 & - 
& 43.7 & 52.2 & 59.3 & 63.1 & 67.2 & 67.8 & 68.7 & - \\


CBST~\cite{zou2018unsupervised} 
& - & 48.8 & - & - & 58.3 & 59.4 & 59.7 & - 
& - & 53.0 & - & - & 66.5 & 69.6 & 71.6 & - \\

GPC~\cite{jiang2021guided} 
& - & - & - & 41.8 & 49.9 & 58.8 & - & 65.8 
& - & - & - & - & - & - & - & - \\

CPS~\cite{chen2021semi} 
& - & 46.7 & - & - & 58.7 & 59.6 & 60.5 & - 
& - & 52.9 & - & - & 66.3 & 70.0 & 72.5 & - \\

\footnotesize{Unal et al}.~\cite{unal2022scribble} 
& - & - & - & 49.9 & 58.7 & 59.1 & - & 68.2 
& - & - & - & - & - & - & - & - \\

LiM3d~\cite{li2023less} 
& - & \underline{58.4} & - & 59.5 & 62.2 & \underline{63.1} & \underline{63.6} & 69.5 
& - & - & - & - & - & - & - & - \\

LaserMix$^{\dagger}$~\cite{kong2023lasermix} 
& \underline{47.3} & 55.5 & \underline{59.2} & \underline{61.7} 
& \underline{62.4} & 62.4 & 62.1 & 63.8
& \underline{51.4} & \underline{58.4} 
& \underline{63.9} & \underline{69.7} 
& \underline{71.6} & \underline{73.7} 
& \underline{73.7} & 75.2 \\
\midrule
\footnotesize{Our method} 
& \textbf{52.2} & \textbf{61.1} 
& \textbf{62.9} & \textbf{62.8} 
& \textbf{63.1} & \textbf{64.1} 
& \textbf{64.3} & - 
& \textbf{58.0} & \textbf{62.9} 
& \textbf{67.2} & \textbf{72.8} 
& \textbf{74.3} & \textbf{76.0} 
& \textbf{76.1} & - \\

$\Delta$ $\uparrow$ 
& \textcolor{brown}{\footnotesize{$\mathbf{+4.9}$}} 
& \textcolor{brown}{\footnotesize{$\mathbf{+5.6}$}} 
& \textcolor{brown}{\footnotesize{$\mathbf{+3.7}$}} 
& \textcolor{brown}{\footnotesize{$\mathbf{+1.1}$}}
& \textcolor{brown}{\footnotesize{$\mathbf{+0.7}$}} 
& \textcolor{brown}{\footnotesize{$\mathbf{+1.7}$}} 
& \textcolor{brown}{\footnotesize{$\mathbf{+2.2}$}} 
& \textcolor{brown}{\footnotesize{$\mathbf{}$}}
& \textcolor{brown}{\footnotesize{$\mathbf{+6.6}$}} 
& \textcolor{brown}{\footnotesize{$\mathbf{+4.5}$}} 
& \textcolor{brown}{\footnotesize{$\mathbf{+3.3}$}} 
& \textcolor{brown}{\footnotesize{$\mathbf{+3.1}$}} 
& \textcolor{brown}{\footnotesize{$\mathbf{+2.7}$}} 
& \textcolor{brown}{\footnotesize{$\mathbf{+2.3}$}} 
& \textcolor{brown}{\footnotesize{$\mathbf{+2.4}$}} 
& \textcolor{brown}{\footnotesize{$\mathbf{}$}}                      \\ \bottomrule
\end{tabular}
\caption{\textbf{Segmentation results (mIoU) on benchmark datasets.} Our method achieves state-of-the-art performance across all labeled ratios. The \textbf{best} and \underline{second best} scores for each ratio are highlighted in \textbf{bold} and \underline{underline}, respectively. $\dagger$ denotes the reproduced result. $\Delta$ indicates an improvement over LaserMix.}
\label{tab:benchmark}
\vspace{-1.5em}
\end{table*}

%% file: sections/4_experiments.tex
\begin{figure*}[]
\begin{center}
\vspace{1.5mm}
\includegraphics[width=\textwidth]{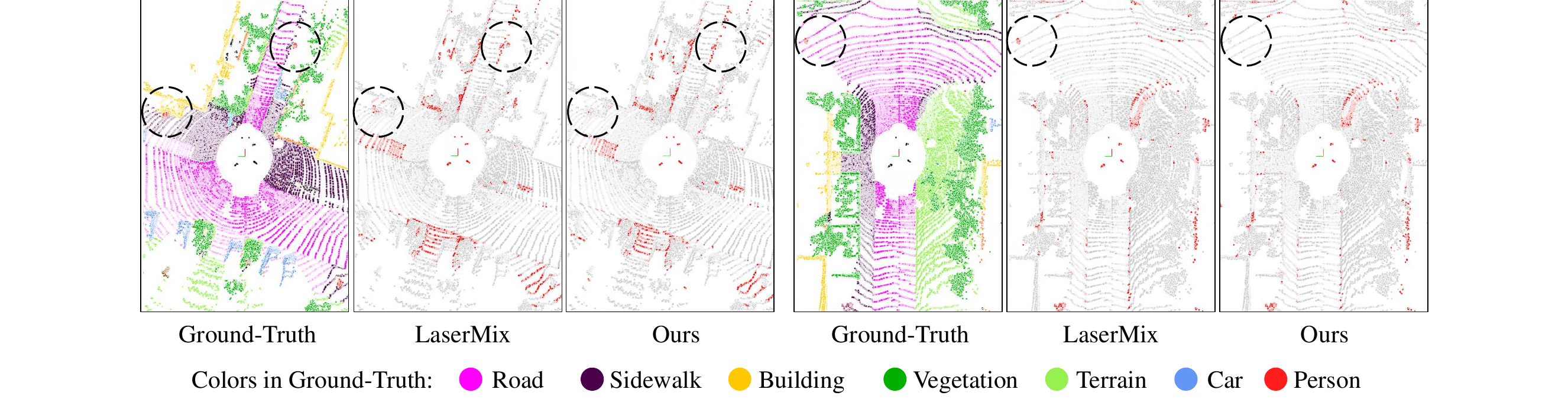}
\end{center}
\vspace{-1em}
\caption{\textbf{Qualitative comparisons between our method and LaserMix.} All samples are visualized from LiDAR bird's-eye view on \textit{val} set of SemanticKITTI~\cite{behley2019semantickitti}. The \textbf{\textcolor{lightgray}{correct}} and \textbf{\textcolor{red}{incorrect}} are painted in \textbf{\textcolor{lightgray}{gray}} and \textbf{\textcolor{red}{red}} to highlight the difference. Dashed circles highlight the misprediction of LaserMix. Best viewed in color.}
\label{fig:qualitative}
\vspace{-1em}
\end{figure*}

\section{Experiment}

\subsection{Settings}
\noindent{\textbf{Datasets.}} We conduct an empirical study on two popular benchmark datasets: SemanticKITTI~\cite{behley2019semantickitti} and nuScenes~\cite{caesar2020nuscenes}. SemanticKITTI contains 19 classes with 19,130 training scans and 4,071 validation scans. nuScenes consists of 16 classes with 29,130 scans for training and 6,019 scans for validation. For the low-budget semi-supervised LiDAR semantic segmentation, we adopt a uniform sampling strategy, selecting 0.5\%, 1\%, 2\%, and 5\% of training scans for labeled data, designating the remainder as unlabeled data. Also, we select 10\%, 20\%, and 50\% of training scans for the normal budget setting. This selection strategy aligns with the conventional semi-supervised settings~\cite{kong2023lasermix, zou2020pseudoseg, chen2021semi}.

\noindent{\textbf{Implementation details.}} We select the voxel-based network, Cylinder3D~\cite{zhou2020cylinder3d}, as the segmentation backbone, configuring the voxel resolution as $[240, 180, 20]$ to accommodate LaserMix~\cite{kong2023lasermix}. Additionally, the hyperparameters for MeanTeacher~\cite{tarvainen2017mean} and LaserMix are adopted based on the settings of LaserMix. To validate the effectiveness of our proposed method, we measure the intersection-over-union (IoU) for each class and reported the mean IoU (mIoU). All experiments were conducted on four NVIDIA A6000 GPUs with 48GB memory using PyTorch.

\subsection{Comparison with SoTA}

Tab.~\ref{tab:benchmark} compares our method with current SoTA methods~\cite{kong2023lasermix,tarvainen2017mean,li2023less,chen2021semi,jiang2021guided,zou2018unsupervised} on the SemanticKITTI and nuScenes datasets. Our method demonstrates clear improvements across all labeled ratios on both datasets, especially under low ratio settings. We see over over 3\%p increases at 0.5\%--2\% labeled ratios. In the normal budget settings (10\%--50\%), our method exhibits moderate improvements of approximately 0.7--2.7\%p. Notably, our method achieves over 64\% and 75\% mIoU on the 20\% ratio in both datasets, outperforming LaserMix's full-label benchmark with only 20\% of labels. This highlights the effectiveness of our PLE and dual-branch architecture in generating accurate pseudo-labels and reducing performance degradation, respectively. Our method establishes new SoTA benchmarks in semi-supervised LiDAR segmentation, surpassing other leading techniques across all labeling ratios.

Fig~\ref{fig:qualitative} presents qualitative examples in a bird's-eye view from SemanticKITTI. Correct predictions are marked in green, while errors are marked in red. Notably, our results exhibit greater accuracy than LaserMix. We specifically emphasize \textit{person} categories with dashed circles, underscoring the superior accuracy of our method in identifying \textit{person} classes compared to LaserMix. This is also supported by the class-wise performance, where \textit{person} and \textit{motorcycle} show 8.4\%p and 44.2\% improvements. These classes have a significant impact on real-world scenarios, in terms of the safety and reliability of autonomous driving.

\input{tables/ple}
\vspace{-0.5em}
\input{tables/ple_classwise}

\begin{figure}[t]
\vspace{-1em}
\centering
\includegraphics[width=\linewidth]{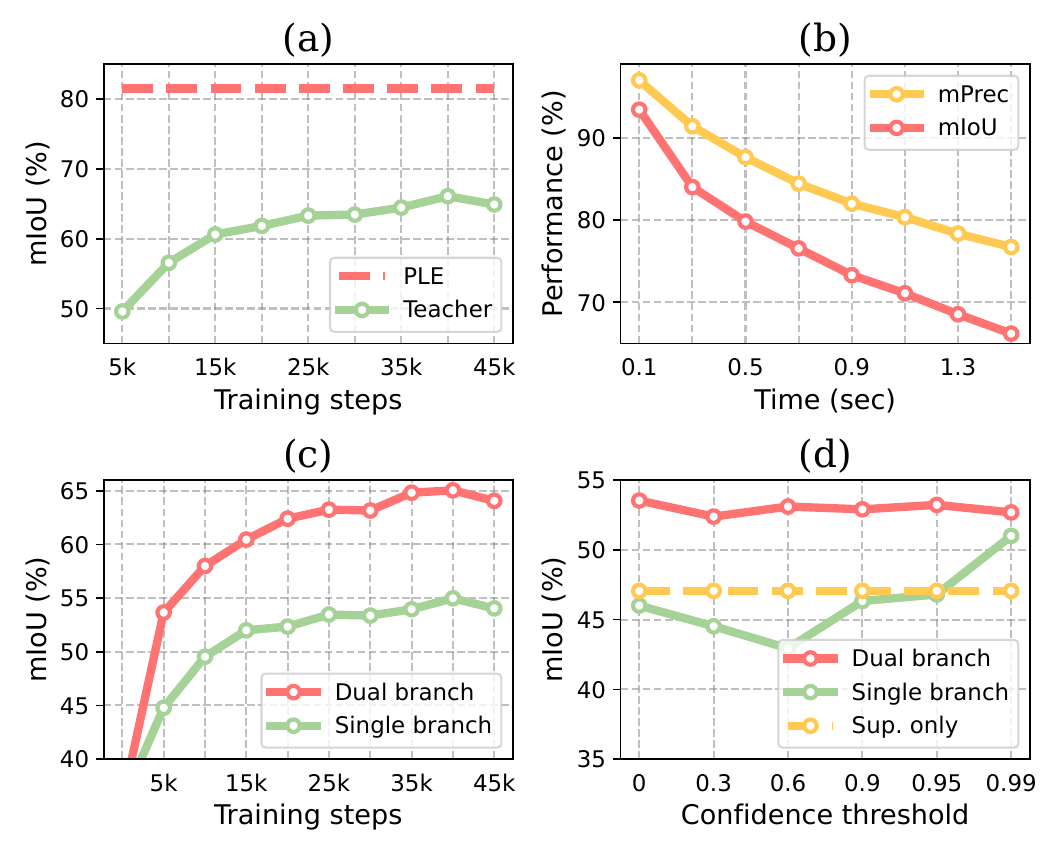}
\vspace{-1.5em}
\caption{\textbf{Ablation studies.} (a) Accuracy of pseudo-labels from the Teacher network and PLE. (b) Accuracy of PLE-labels over time intervals. (c) Accuracy of pseudo-labels during training. (d) Training results according to confidence threshold. All results are from a 1\%  ratio of the SemanticKITTI.}
\label{fig:ablation}
\vspace{-2em}
\end{figure}

\subsection{Ablation Study}

\noindent{\textbf{Accuracy of PLE labels.}} In Tab.~\ref{tab:tab_ple}, we detail the impact of applying PLE on the statistics of training data and the accuracy of PLE labels. Using PLE, we search labeled data within a $\pm$ 1-second interval around each unlabeled scan. With the 10Hz acquisition rate of SemanticKITTI, our method assigns pseudo-labels up to 20 unlabeled frames around each labeled frame, increasing the size of the labeled set by about 19 times. At a 5\% labeled ratio, our PLE generates over half of the total LiDAR scans. Furthermore, the PLE labels are highly accurate, achieving a mIoU surpassing 79\% and outperforming state-of-the-art fully supervised models with approximately 70\% mIoU~\cite{zhou2020cylinder3d,hou2022point}. Notably, the progressive version of PLE yields more accurate labels compared to the naive version, particularly with significant improvements in dynamic classes. As shown in Tab.~\ref{tab:ple_classwise}, there is a remarkable increase of 5.2\%--19.5\% in in dynamic classes, highlighting the effectiveness of using intermediate PLE labels during the PLE process instead of solely relying on direct references from labeled samples. This strategy minimizes errors in capturing the movement of dynamic objects, proving its effectiveness in handling the complexities of dynamic classes.

\input{tables/component}

\noindent{\textbf{Pseudo-labels from the teacher network.}} We compare the performance between the PLE labels and the pseudo-labels generated by the Teacher network in Fig.~\ref{fig:ablation}(a). During training, the quality of the pseudo-labels from the Teacher network initially improves but eventually saturates and even slightly declines. In contrast, PLE labels consistently present remarkably higher accuracy (over 10\%p) than the pseudo-labels from the Teacher network. This result indicates that PLE labels lead to significant improvement in our semi-supervised training.

\noindent{\textbf{PLE labels according to time intervals.}} Fig.~\ref{fig:ablation}(b) depicts the accuracy of PLE labels on SemanticKITTI relative to varying time intervals (frame difference), achieved by selecting adjacent frames at fixed time differences. The quality of PLE labels deteriorates with increasing time gaps between scans. This is due to reduced scan overlap over time, as well as dynamic changes in object movement and occlusion within the scene. Throughout the experiments, we select adjacent scans within a 1-second interval, ensuring reasonable temporal proximity and performance.

\noindent{\textbf{Effectiveness of dual-branch.}} Fig.~\ref{fig:ablation}(c) compares the quality of pseudo-labels in single-branch and dual-branch settings during training. The dual-branch consistently enhances the quality of pseudo-labels, particularly presenting strong performance from the early training stages. Also, we compare the segmentation performance of the dual-branch structure with the single branch according to confidence threshold values in Fig.~\ref{fig:ablation}(d). The single branch network occasionally underperforms compared to using only labeled data. In contrast, the dual-branch is robust to confidence thresholds and consistently outperforms the single-branch at all thresholds. This improvement is because the dual-branch breaks the vicious training cycle where noisy pseudo-labels affect the training process of the Teacher network and vice versa.

\noindent{\textbf{Effect of each component.}} In Tab.~\ref{tab:component}, we evaluate the effectiveness of individual components of our approach at various labeled ratios (0.5\%, 1\%, 2\%, and 5\%). The results demonstrate a notable performance gain by employing PLE and the dual-branch, respectively. Specifically, PLE enhances accuracy by 1.8--4.7\%p over the MeanTeacher, with the progressive reference of the PLE process further boosting performance by 0.3-–1.9\%p. This improvement is attributed to the superior accuracy of PLE labels compared to those from the teacher network, as detailed in Fig.~\ref{fig:ablation}(a). Moreover, the dual-branch consistently improves the performance of the MeanTeacher. Unlike the original MeanTeacher, which uses a single branch, our modification utilizing a dual-branch eliminates noisy training signals and instead focuses on clean labels for training the teacher's C-branch. This has greater efficacy at lower ratios (e.g., 0.5\% ratio translates to 99.5\% of noisy labels for training the MeanTeacher), highlighting its suitability for low-budget scenarios. Finally, when we apply PLE and the dual-branch simultaneously, we achieve further performance gain. This is because each component is mutually exclusive when it is applied to the unlabeled data. PLE generates accurate pseudo-labels for the nearby unlabeled scans of the labeled data, while the dual-branch separates the remaining unlabeled data from the C-branch. This highlights the complementary effects of PLE and the dual-branch, thereby establishing the effectiveness of each module in our method.

\input{tables/tab_extension}

\noindent{\textbf{Representation and backbones.}} To demonstrate the scalability of our method, we evaluate it across other representation and backbones. Specifically, we incorporate the widely used CENet~\cite{cheng2022cenet} in the range view, and introduce MinkUNet~\cite{choy20194d} and SPVCNN~\cite{tang2020searching}. As depicted in Tab.~\ref{tab:extension}, our approach consistently significantly enhances performance consistently across different types of representations and backbones. Notably, in the range view, we observed substantial improvements ranging from 2.1--6.3\%p. Similarly, with MinkUNet18 and SPVCNN, performance improvements ranged from 1.5--4.4\%p. The substantial enhancements achieved through the use of PLE and a dual-branch architecture underscore their versatility and applicability across different representations and backbone architectures.

%% file: tables/ple.tex
\begin{table}[t!]
\centering

{\small
\begin{tabularx}{.48\textwidth}{>{\raggedright\arraybackslash}Xrrrr}
\toprule
\multicolumn{1}{c}{} & \multicolumn{1}{c}{0.5\%} & \multicolumn{1}{c}{1\%} & \multicolumn{1}{c}{2\%} & \multicolumn{1}{c}{5\%} \\ \midrule
\# GT labels & 95 & 191 & 382 & 956 \\
\# PLE labels & 1,805 & 3,404 & 6,247 & 11,490 \\
\# Unlabeled data & 17,230 & 15,535 & 12,501 & 6,684 \\
\rowcolor{lightgray} \multicolumn{5}{c}{\textit{Naive PLE}} \\
mIoU (\%) & 79.5 & 79.9 & 79.8 & 81.1 \\
mPrecision (\%) & 86.5 & 87.6 & 87.4 & 88.4 \\
\rowcolor{lightgray} \multicolumn{5}{c}{\textit{Progressive PLE}} \\
mIoU (\%) & 81.1 & 81.5 & 81.7 & 83.0 \\
mPrecision (\%) & 88.7 & 88.9 & 89.3 & 89.9 \\ \bottomrule
\end{tabularx}
}
\caption{\textbf{Statistics changes of training data and accuracy of PLE labels on SemanticKITTI~\cite{behley2019semantickitti}.} IoU and Precision are evaluated on each ratio of PLE labels.}
\label{tab:tab_ple}
\vspace{-0.5em}
\end{table}

%% file: tables/ple_classwise.tex
\begin{table}[]
\centering
\begin{tabular}{>{\raggedright\arraybackslash}lcccccccc}
\toprule
            & mean & car  & bicy & moto & ped  & b.cyc & m.cyc \\ \midrule
Naive       & 79.9 & 87.1 & 61.2 & 78.2 & 82.1 & 34.7  & 13.8  \\
Prg. & 81.5 & 87.7 & 66.6 & 83.4 & 83.5 & 54.2  & 20.6  \\ \bottomrule
\end{tabular}
\caption{\textbf{Accuracy (IoU) of dynamic classes in PLE labels.} Prg. refers to progressive version of PLE.}
\label{tab:ple_classwise}
\vspace{-2em}
\end{table}

%% file: tables/component.tex
\begin{table}
\centering

\vspace{1.5mm}
{\small
\begin{tabular}[width=\linewidth]{>{\raggedright\arraybackslash}ccccccc}
\toprule
    PLE & PRG & DB & 0.5\% & 1.0\% & 2.0\% & 5.0\% \\\midrule

               &            &            & 42.2           & 51.0          & 54.6          & 55.6 \\
    \checkmark &            &            & 45.2           & 55.7          & 56.9          & 57.4 \\
    \checkmark & \checkmark &            & 47.1           & 56.0          & 57.5          & 59.3 \\ 
               &            & \checkmark & 47.9           & 52.9          & 55.7          & 57.9 \\ 
    \checkmark & \checkmark & \checkmark & \textbf{48.2}  & \textbf{56.6} & \textbf{58.9} & \textbf{59.9} \\
\bottomrule
\end{tabular}
}
\caption{\textbf{Ablation study on each component.} The accuracy (mIoU) is evaluated on the \textit{val} set in SemanticKITTI. MeanTeacher is used as a baseline. DB and PRG refer to dual-branch and the progressive PLE. The \textbf{best score} for each ratio is highlighted in \textbf{bold}.}
\label{tab:component}
\vspace{-2.5em}
\end{table}

%% file: tables/tab_extension.tex
\begin{table}[]
\centering

\vspace{1.5mm}
{\small
\begin{tabular}{>{\raggedright\arraybackslash}lcccc}
\toprule
& 0.5\% & 1\%  & 2\%  & 5\%  \\ \midrule
\rowcolor{lightgray} \multicolumn{5}{c}{\textit{Range View}} \\
CENet & 42.3  & 45.2 & 52.2 & 55.1 \\
+ Ours & 46.2 \textcolor{brown}{\footnotesize{$\mathbf{+3.9}$}} & 51.5 \textcolor{brown}{\footnotesize{$\mathbf{+6.3}$}} & 54.3 \textcolor{brown}{\footnotesize{$\mathbf{+2.1}$}} & 58.1 \textcolor{brown}{\footnotesize{$\mathbf{+3.0}$}} \\ \midrule

\rowcolor{lightgray} \multicolumn{5}{c}{\textit{Voxel View}} \\
Minku18 & 51.9  & 57.3 & 57.6 & 59.0 \\
+ Ours & 56.1 \textcolor{brown}{\footnotesize{$\mathbf{+4.2}$}}  & 60.1 \textcolor{brown}{\footnotesize{$\mathbf{+2.8}$}} & 60.9 \textcolor{brown}{\footnotesize{$\mathbf{+3.3}$}} & 63.4 \textcolor{brown}{\footnotesize{$\mathbf{+4.4}$}} \\
SPVCNN & 51.5  & 56.9 & 57.2 & 59.3 \\
+ Ours & 54.9 \textcolor{brown}{\footnotesize{$\mathbf{+3.4}$}}  & 60.7 \textcolor{brown}{\footnotesize{$\mathbf{+3.8}$}} & 60.4 \textcolor{brown}{\footnotesize{$\mathbf{+3.2}$}} & 60.8 \textcolor{brown}{\footnotesize{$\mathbf{+1.5}$}} \\ \bottomrule
\end{tabular}
}
\caption{\textbf{Ablation study on other representation and backbones.} The accuracy (mIoU) is evaluated on the \textit{val} set in SemanticKITTI. MT and DB refer to MeanTeacher~\cite{tarvainen2017mean} and dual-branch, respectively.}
\label{tab:extension}
\vspace{-1.5em}
\end{table}

%% file: sections/5_conclusion.tex
\section{Conclusion}

Our work tackles key challenges in semi-supervised LiDAR segmentation, focusing on low-budget scenarios. We introduce a pseudo-labeling approach, Proximity-based Label Estimation (PLE), leveraging LiDAR's spatio-temporal prior to enhances the quality of pseudo-labels. Our dual-branch structure overcomes the labeled-unlabeled data imbalance, ensuring strong performance. Through extensive experiments, we demonstrate our method's efficacy, setting new state-of-the-art results on SemanticKITTI and nuScenes and surpassing the fully supervised counterparts. Based on its effectiveness and simplicity, our approach significantly boosts scalable LiDAR perception, expediting its application in real-world scenarios.

